\def\year{2016}
\documentclass[letterpaper]{article}
\usepackage{aaai16}
\usepackage{times}
\usepackage{helvet}
\usepackage{courier}
\usepackage{color}
\usepackage{url}
\usepackage{danudefs}
\usepackage{graphicx}
\usepackage{epstopdf}

\newcommand{\citet}[1]{\citeauthor{#1} \shortcite{#1}}
\newcommand{\citep}{\cite}

\usepackage{algorithmic}
\usepackage{algorithm}

\begin{document}
%
\title{Joint Word  Representation Learning using a Corpus and a Semantic Lexicon}


\author{Danushka Bollegala\thanks{Japan Science and Technology Agency, ERATO, Kawarabayashi Large Graph Project} \\ University of Liverpool \And
 Alsuhaibani Mohammed \\ University of Liverpool \And
Takanori Maehara\footnotemark[1] \\
Shizuoka University  \And
Ken-ichi Kawarabayashi\footnotemark[1] \\
National Institute of Informatics}

\maketitle
\begin{abstract}
Methods for learning word representations using large text corpora have received much attention lately due to their impressive performance
in numerous natural language processing (NLP) tasks such as, semantic similarity measurement, and word analogy detection.
Despite their success, these data-driven word representation learning methods do not consider
the rich semantic relational structure between words in a co-occurring context. 
On the other hand, already much manual effort has gone into the construction of semantic lexicons such as the WordNet
that represent the meanings of words by defining the various relationships that exist among the words in a language.
We consider the question, \emph{can we improve the word representations learnt using a corpora by integrating the
knowledge from semantic lexicons?}. For this purpose, we propose a \emph{joint} word representation learning method that simultaneously predicts
the co-occurrences of two words in a sentence subject to the relational constrains given by the semantic lexicon.
We use relations that exist between words in the lexicon to regularize the word representations learnt from the corpus.
Our proposed method statistically significantly outperforms previously proposed methods for incorporating semantic lexicons into word
representations on several benchmark datasets for semantic similarity and word analogy.
\end{abstract}

\section{Introduction}
\label{sec:intro}

Learning representations for words is a fundamental step in various NLP tasks. If we can accurately represent the meanings of words
using some linear algebraic structure such as vectors, we can use those word-level semantic representations to compute
representations for larger lexical units such as phrases, sentences, 
or texts~\cite{socher-EtAl:2012:EMNLP-CoNLL,Le:ICML:2014}.
Moreover, by using word representations as features in downstream NLP applications, 
significant improvements in performance have been obtained~\cite{Turian:ACL:2010,Bollegala:ACL:2015,Collobert:2011}.
Numerous approaches for learning word representations from large text corpora have been proposed, such as 
\emph{counting-based} methods~\cite{Turney:JAIR:2010} that follow the distributional hypothesis and use contexts
 of a word to represent that word, and \emph{prediction-based} methods~\cite{Mikolov:NIPS:2013} 
 that learn word representations by predicting the occurrences 
 of a word in a given context~\cite{baroni-dinu-kruszewski:2014:P14-1}.

Complementary to the corpus-based data-driven approaches for learning word representations, significant manual effort has
already been invested in creating semantic lexicons such as the WordNet~\cite{WordNet}. 
Semantic lexicons explicitly define the meaning of words by
specifying the relations that exist between words, such as synonymy, hypernymy, or meronymy. 
Although it is attractive to learn word representation purely from a corpus in an unsupervised fashion because it obviates the need
for manual data annotation, there exist several limitations to this corpus-only approach where a semantic lexicon could help to overcome.
First, corpus-based approaches operate on surface-level word co-occurrences, and ignore the rich semantic relations that exist
between the two words that co-occur in the corpus. 
Second, unlike in a semantic lexicon where a word is grouped with the other words with similar senses (e.g., WordNet synsets),
occurrences of a word in a corpus can be ambiguous. Third, the corpus might not be sufficiently large to obtain reliable
word co-occurrence counts, which is problematic when learning representations for rare words.

On the other hand, purely using a semantic lexicon to learn word representations~\cite{Bollegala:AAAI:2015} can also be problematic.
First, unlike in a corpus where we can observe numerous co-occurrences between two words in different contexts,
in a semantic lexicon we often have only a limited number of entries for a  particular word. 
Therefore, it is difficult to accurately estimate the strength of the relationship between two words using only a semantic lexicon. 
Second, a corpus is likely to include neologisms or novel, creative uses of existing words. 
Because most semantic lexicons are maintained manually on a periodical basis, such trends will not be readily reflected in the semantic lexicon.
Considering the weaknesses of methods that use either only a corpus or only a semantic lexicon, it is a natural motivation for us to
explore hybrid approaches.

To illustrate how a semantic lexicon can potentially assist the corpus-based word representation learning process,
let us consider the sentence ``\emph{I like both cats and dogs}''. 
If this was the only sentence in the corpus where the three words \emph{cat}, \emph{dog}, and \emph{like} occur, 
we would learn word representations for those three words 
that predict \emph{cat} to be equally similar to \emph{like} as to \emph{dog} because,
there is exactly one co-occurrence between all three pairs generated from those three words.
However, a semantic lexicon would list \emph{cat} and \emph{dog} as hyponyms of \emph{pet}, but not of \emph{like}.
Therefore, by incorporating such constraints from a semantic lexicon
 into the word representation learning process, we can potentially overcome this problem.
 
We propose a method to learn word representations using both a corpus and a semantic lexicon in a \emph{joint} manner.
Initially, all words representations are randomly initialized with fixed, low-dimensional, real-valued vectors, which are subsequently updated to
predict the co-occurrences between two words in a corpus. 
We use a regularized version of the global co-occurrence prediction approach proposed
by~Pennington et al.~\shortcite{Pennington:EMNLP:2014} as our objective function.
We use the semantic lexicon to construct a regularizer that enforces two words that are in a particular semantic relationship in the lexicon 
to have similar word representations.
Unlike \emph{retrofitting}~\cite{faruqui-EtAl:2015:NAACL-HLT}, 
which fine-tunes pre-trained word representations in a post-processing step, our method
jointly learns both from the corpus as well as from the semantic lexicon, thereby benefitting from the knowledge in the 
semantic lexicon during the word representation learning stage.

In our experiments, we use seven relation types found in the WordNet, and compare the word representations
learnt by the proposed method.
Specifically, we evaluate the learnt word representations on two standard tasks:
semantic similarity prediction~\cite{Bollegala:NAACL:2007}, and word analogy prediction~\cite{Duc:AAAI:2011}.
On both those tasks, our proposed method statistically significantly outperforms all 
previously proposed methods for learning word representations using a semantic lexicon and a corpus.
The performance of the proposed method is stable across a wide-range of vector dimensionalities.
Furthermore, experiments conducted using different sized corpora show that the benefit of incorporating a semantic lexicon
is more prominent for smaller corpora.

\section{Related Work}
\label{sec:related}



Learning word representations using large text corpora has received a renewed interest recently due to the
impressive performance gains obtained in downstream NLP applications using the word representations as features~\cite{Collobert:2011,Turian:ACL:2010}.
Continuous bag-of-words (CBOW)  and skip-gram (SG) methods proposed by~Mikolov et al.~\cite{Milkov:2013}
use the local co-occurrences of a target word and other words in its context for learning word representations.
Specifically, CBOW predicts a target word given its context, whereas SG predicts the context given the target word.
Global vector prediction (GloVe)~\cite{Pennington:EMNLP:2014} on the other hand first builds a word co-occurrence
matrix and predicts the total co-occurrences between a target word and a context word.
Unlike SG and CBOW, GloVe does not require negative training instances, and is less likely to affect from random local
co-occurrences because it operates on global counts. 
However, all of the above mentioned methods are limited to using only a corpus, and
the research on using semantic lexicons in the word representation learning process has been limited.

\citet{Yu:ACL:2014} proposed the \emph{relation constrained model} (RCM) where they used word similarity information to improve the word representations learnt using CBOW. Specifically, RCM assigns high probabilities to words that are listed as similar in the lexicon.
Although we share the same motivation as \citet{Yu:ACL:2014} for jointly learning word representations from a corpus and
a semantic lexicon, our method differs from theirs in several aspects. 
First, unlike in RCM where only synonymy is considered, we use different types of semantic relations in our model. 
As we show later in Section~\ref{sec:exp}, besides synonymy, numerous other semantic relation types are useful for different tasks.
Second, unlike the CBOW objective used in RCM that considers only local co-occurrences, we use
global co-occurrences over the entire corpus. This approach has several benefits over the CBOW method, such as
we are not required to normalize over the entire vocabulary to compute conditional probabilities, which is computationally
costly for large vocabularies. Moreover, we do not require pseudo negative training instances. Instead, the number of
co-occurrences between two words is predicted using the inner-product between the corresponding word representations.
Indeed, \citet{Pennington:EMNLP:2014} show that we can learn superior word representations by predicting global co-occurrences
instead of local co-occurrences. 

\citet{Xu:2014} proposed RC-NET that uses both relational (R-NET) and categorical (C-NET) information in a knowledge base (KB)
jointly with the skip-gram objective for learning word representations.
They represent both words and relations in the same embedded space.
Specifically, given a relational tuple $(h,r,t)$ where a semantic relation $r$ exists in the KB between two words $h$ and $t$,
they enforce the constraint that the vector sum $(\vec{h} + \vec{r})$ of the representations for $h$ and $r$ must be similar to
$\vec{t}$ for words $t$ that have the relation $r$ with $h$. Similar to RCM, RC-NET is limited to using local co-occurrence counts.

In contrast to the joint learning methods discussed above,
\citet{faruqui-EtAl:2015:NAACL-HLT} proposed \emph{retrofitting}, a post-processing method that fits pre-trained word representations for
a given semantic lexicon. 
The modular approach of retrofitting is attractive because it can be used to fit arbitrary pre-trained word representations
to an arbitrary semantic lexicon, without having to retrain the word representations. 
 \citet{johansson-nietopina:2015:NAACL-HLT} proposed a method to embed 
a semantic network consisting of linked word senses into a continuous-vector word space. Similar to retrofitting, their method takes
pre-trained word vectors and computes sense vectors over a given semantic network.
However, a disadvantage of such an approach is that we cannot use the rich information in the semantic lexicon when we
learn the word representations from the corpus. Moreover,  incompatibilities between the corpus and the lexicon, such as
the differences in word senses, and missing terms must be carefully considered.
We experimentally show that our joint learning approach outperforms the post-processing approach used in retrofitting.

\citet{iacobacci-pilehvar-navigli:2015:ACL-IJCNLP} used BabelNet and consider words that are connected to a source word
in BabelNet to overcome the difficulties when measuring the similarity between rare words. However, they do not consider the semantic
relations between words and only consider words that are listed as related in the BabelNet, which encompasses multiple semantic relations.
\citet{Bollegala:AAAI:2015} proposed a method for learning word representations from a relational graph, where they represent 
words and relations respectively by vectors and matrices. Their method can be applied on either a manually created relational graph,
or an automatically extracted one from data. However, during training they use only the relational graph and do not use the corpus.

\section{Learning Word Representations}
\label{sec:method}

Given a corpus $\cC$, and a semantic lexicon $\cS$, we describe a method for learning
word representations $\vec{w}_i \in \R^d$ for words $w_i$ in the corpus. 
We use the boldface $\vec{w}_i$ to denote the word (vector) representation of
the $i$-th word $w_i$, and the vocabulary (i.e., the set of all words in the corpus) is denoted by $\cV$.
The dimensionality $d$ of the vector representation is a hyperparameter of the proposed method that must be
specified by the user in advance.  Any semantic lexicon that specifies the semantic relations that exist between words could be used as $\cS$,
such as the WordNet~\citep{WordNet}, FrameNet~\citep{baker-fillmore-lowe:1998:ACLCOLING}, or the Paraphrase Database~\citep{ganitkevitch-vandurme-callisonburch:2013:NAACL-HLT}. In particular, we do not assume any structural properties unique to a particular semantic lexicon. In the experiments described in this paper we use the WordNet as the semantic lexicon.

Following~\citet{Pennington:EMNLP:2014}, first we create a co-occurrence matrix $\mat{X}$ in which
words that we would like to learn representations for (\emph{target words}) are arranged in rows of $\mat{X}$, whereas words
that co-occur with the target words in some contexts (\emph{context words}) are arranged in columns of $\mat{X}$.
The $(i,j)$-th element $X_{ij}$ of $\mat{X}$ is set to the total co-occurrences of $i$ and $j$ in the corpus.
Following the recommendations in prior work on word representation learning~\citep{Levy:TACL:2015},
we set the context window to the $10$ tokens preceding and succeeding a word in a sentence.
We then extract unigrams from the co-occurrence windows
 as the corresponding context words. We down-weight distant (and potentially noisy) co-occurrences
 using the reciprocal $1/l$ of the distance in tokens $l$ between the two words that co-occur.

A word $w_i$ is assigned two vectors $\vec{w}_i$ and $\tilde{\vec{w}}_i$ denoting whether $w_i$ is respectively
the target of the prediction  (corresponding to the rows of $\mat{X}$),
or in the context of another word (corresponding to the columns of $\mat{X}$).
The GloVe objective can then be written as:
\begin{equation}
\small
\label{eq:corp}
J_{\cC} = \frac{1}{2} \sum_{i \in \cV} \sum_{j \in \cV} f(X_{ij}) {\left(\vec{w}_i\T\tilde{\vec{w}}_j +b_i + \tilde{b}_j  - \log(X_{ij}) \right)}^2
\end{equation}
Here, $b_i$ and $\tilde{b}_j$ are real-valued scalar bias terms that adjust for
the difference between the inner-product and the logarithm of
the co-occurrence counts.
The function $f$ discounts the co-occurrences between frequent words and is given by:
\begin{equation}
\small
f(t) = \begin{cases} (t/t_{\max})^\alpha & \text{if $t < t_{\max}$} \\
				1 & \text{otherwise}
	\end{cases}
\end{equation}
Following~\cite{Pennington:EMNLP:2014}, we set $\alpha = 0.75$ and $t_{\max} = 100$ in our experiments.
The objective function defined by \eqref{eq:corp} encourages the learning of word representations that demonstrate the
desirable property that vector difference between the word embeddings for two words represents the semantic relations
that exist between those two words. For example, \citet{Mikolov:NAACL:2013} observed that 
the difference between the word embeddings for the words \emph{king} and \emph{man} when added to the word embedding
for the word \emph{woman} yields a vector similar to that of \emph{queen}.

Unfortunately, the objective function given by \eqref{eq:corp} does not capture the semantic relations that exist between $w_i$ and $w_j$
as specified in the lexicon $\cS$. Consequently, it considers all co-occurrences equally and is likely to encounter problems when the
co-occurrences are rare. 
To overcome this problem we propose a regularizer, $J_{\cS}$,  by considering the three-way co-occurrence
among words $w_i$, $w_j$, and a semantic relation $R$ that exists between the target word $w_i$ and 
one of its context words $w_j$ in the lexicon as follows:
\begin{equation}
\small
\label{eq:onto}
J_{\cS} = \frac{1}{2} \sum_{i \in \cV}  \sum_{j \in \cV} R(i,j) \left(\vec{w}_i - \tilde{\vec{w}}_j \right)^2 
\end{equation}
Here, $R(i,j)$ is a binary function that returns $1$ if  the semantic relation $R$ exists between the words $w_i$ and $w_j$
in the lexicon, and $0$ otherwise. 
In general, semantic relations are asymmetric. Thus, we have $R(i,j) \neq R(j,i)$.
Experimentally, we consider both symmetric relation types, such as synonymy and antonymy, as well as
asymmetric relation types, such as hypernymy and meronymy. 
The regularizer given by \eqref{eq:onto} enforces the constraint that the words that are connected by a semantic relation $R$ in the
lexicon must have similar word representations.

We would like to learn target and context word representations  $\vec{w}_i$, $\tilde{\vec{w}}_j$ 
that simultaneously minimize both \eqref{eq:corp} and \eqref{eq:onto}.
Therefore, we formulate the joint objective as a minimization problem as follows:
\begin{equation}
\small
\label{eq:overall}
J = J_{\cC} + \lambda J_{\cS}
\end{equation}
Here, $\lambda \in \R^+$ is a non-negative real-valued regularization coefficient that determines
the influence imparted by the semantic lexicon on the word representations learnt from the corpus. 
We use development data to estimate the optimal value of $\lambda$ as described later in Section~\ref{sec:exp}.

The overall objective function given by \eqref{eq:overall} is non-convex w.r.t. to the four variables $\vec{w}_i$, $\tilde{\vec{w}}_j$, $b_i$, and
$\tilde{b}_j$. However, if we fix three of those variables, then \eqref{eq:overall} becomes convex in the remaining one variable.
We use an alternative optimization approach where we first randomly initialize all the parameters, and then cycle through the set of variables
in a pre-determined order updating one variable at a time while keeping the other variables fixed.

The derivatives of the objective function w.r.t. the variables are given as follows:
{\small
\begin{align}
\label{eq:part:wi}
\frac{\partial J}{\partial \vec{w}_i} &=&   \sum_j f(X_{ij}) \tilde{\vec{w}}_j \left(\vec{w}_i\T\tilde{\vec{w}}_j +b_i + \tilde{b}_j  - \log(X_{ij}) \right) \nonumber \\
							   & & + \lambda \sum_j R(i,j) (\vec{w}_i - \tilde{\vec{w}}_j)	\\	
\label{eq:part:bi}					   
\frac{\partial J}{\partial b_i} &=&   \sum_j f(X_{ij}) \left(\vec{w}_i\T\tilde{\vec{w}}_j +b_i + \tilde{b}_j  - \log(X_{ij}) \right) 	\\	
\label{eq:part:wj}				   
\frac{\partial J}{\partial \tilde{\vec{w}}_j} &=&  \sum_i f(X_{ij}) \vec{w}_i \left(\vec{w}_i\T\tilde{\vec{w}}_j +b_i + \tilde{b}_j  - \log(X_{ij}) \right) \nonumber \\
 							   & &	- \lambda \sum_j R(i,j) (\vec{w}_i - \tilde{\vec{w}}_j)	\\	
\label{eq:part:bj}					   
\frac{\partial J}{\partial \tilde{b}_j} &=&   \sum_i f(X_{ij}) \left(\vec{w}_i\T\tilde{\vec{w}}_j +b_i + \tilde{b}_j  - \log(X_{ij}) \right) 
\end{align}
}
We use stochastic gradient descent (SGD) with learning rate scheduled by
AdaGrad~\citep{Duchi:JMLR:2011} as the optimization method.
The overall algorithm for learning word embeddings is listed in Algorithm~\ref{algo:ontorep}.

We used the ukWaC\footnote{\url{http://wacky.sslmit.unibo.it}} as the corpus. 
It has ca. 2 billion tokens and have been used for learning word embeddings in prior work.
We initialize word embeddings by randomly sampling each dimension from the uniform distribution in the range $[-1,+1]$.
We set the initial learning rate in AdaGrad to $0.01$ in our experiments. 
We observed that $T = 20$ iterations is sufficient for the proposed method to converge to a solution.

Building the co-occurrence matrix $\mat{X}$ is an essential pre-processing step for the proposed method.
Because the co-occurrences between rare words will also be rare, we can first count the frequency of each word and
drop words that have total frequency less than a pre-defined threshold to manage the memory requirements of the
co-occurrence matrix. In our experiments, we dropped words that are less than $20$ times in the entire corpus
when building the co-occurrence matrix. 
For storing larger co-occurrence matrices we can use distributed hash tables and sparse representations.

The \emph{for-loop} in Line 3 of Algorithm~\ref{algo:ontorep} iterates over the non-zero
elements in $\mat{X}$. If the number of non-zero elements in $\mat{X}$ is $n$, the overall time complexity
of Algorithm~\ref{algo:ontorep} can be estimated as $\O(|\cV|dTn)$, where $|\cV|$ denotes the number of words
in the vocabulary. 
Typically, the global co-occurrence matrix is highly sparse, containing less than $0.03\%$ of non-zero entries.
It takes under 50 mins. to learn $300$ dimensional word representations for $|\cV| = 434,826$ words 
($n = 58,494,880$) from the ukWaC corpus
on a Xeon 2.9GHz 32 core 512GB RAM machine. 
The source code and data for the proposed method is publicly available\footnote{\url{https://github.com/Bollegala/jointreps}}.

\begin{algorithm}[t]     
\small  
\caption{Jointly learning word representations using a corpus and a semantic lexicon.}        
\label{algo:ontorep}                         
\begin{algorithmic}[1]         
\REQUIRE Word co-occurrence matrix $\mat{X}$ specifying the co-occurrences between words in the corpus $\cC$, 
		relation function $R$ specifying the semantic relations between words in the lexicon $\cS$, 
		dimensionality $d$ of the word embeddings, and the maximum number of iterations $T$.
\ENSURE Embeddings $\vec{w}_i, \tilde{\vec{w}}_j \in \R^d$, of all words $i, j$ in the vocabulary $\cV$.
\medskip
\STATE Initialize word vectors $\vec{w}_i, \tilde{\vec{w}}_j \in \R^d$ randomly. \label{line:init}
\FOR{ $t = 1$ \TO $T$ }
	\FOR{$(i,j) \in \mat{X}$}
		\STATE Use \eqref{eq:part:wi} to update $\vec{w}_i$
		\STATE Use \eqref{eq:part:bi} to update $b_i$
		\STATE Use \eqref{eq:part:wj} to update $\tilde{\vec{w}}_j$
		\STATE Use \eqref{eq:part:bj} to update $b_j$
	\ENDFOR 
\ENDFOR 
\RETURN $\vec{w}_i, \tilde{\vec{w}}_j$ for all words $i,j \in \cV$.
\end{algorithmic}
\end{algorithm}

\section{Experiments and Results}
\label{sec:exp}

We evaluate the proposed method on two standard tasks: predicting the semantic similarity between two words,
and predicting proportional analogies consisting of two pairs of words.
For the similarity prediction task, we use the following benchmark datasets: Rubenstein-Goodenough (\textbf{RG}, 65 word-pairs)~\citep{RG},
Miller-Charles (\textbf{MC}, 30 word-pairs)~\citep{MC}, rare words dataset (\textbf{RW}, 2034 word-pairs)~\citep{Luong:CoNLL:2013},
Stanford's contextual word similarities (\textbf{SCWS}, 2023 word-pairs)~\citep{Huang:ACL:2012}, 
and the \textbf{MEN} test collection (3000 word-pairs)~\citep{MEN}.
Each word-pair in those benchmark datasets has a manually assigned similarity score,
 which we consider as the gold standard rating for semantic similarity.  

For each word $w_i$, the proposed method learns a target representation $\vec{w}_i$, and a context representation $\tilde{\vec{w}}_i$.
\cite{Levy:TACL:2015} show that the addition of the two vectors, $\vec{w}_i + \tilde{\vec{w}}_i$, gives a better 
representation for the word $w_i$. In particular, when we measure the cosine similarity between two words using
their word representations, this additive approach considers both first and second-order similarities between the two words.
\cite{Pennington:EMNLP:2014} originally motivated this additive operation as an ensemble method.
Following these prior recommendations, we add the target and context representations to create the final representation for a word.
The remainder of the experiments in the paper use those word representations.

Next, we compute the cosine similarity between the two corresponding
embeddings of the words. Following the standard approach for evaluating using the above-mentioned benchmarks,
we measure Spearman correlation coefficient between gold standard ratings and the predicted similarity scores.
We use the Fisher transformation to test for the statistical significance of the correlations.
Table~\ref{tbl:overall} shows the Spearman correlation coefficients  on the five similarity benchmarks,
where high values indicate a better agreement with the human notion of semantic similarity.

\begin{table*}[t!]
\small
\centering
\caption{Performance of the proposed method with different semantic relation types.}
\label{tbl:overall}
\begin{tabular}{|l|c|c|c|c|c|c|c|c|c|} \hline
Method & RG & MC & RW & SCWS & MEN & sem & syn & total & SemEval \\ \hline \hline
corpus only  & 0.7523 & 0.6398 & 0.2708 & 0.460 & 0.6933 & 61.49 & 66.00 & 63.95 & 37.98 \\
Synonyms & \textbf{0.7866} & \textbf{0.7019} & 0.2731 & \textbf{0.4705} & \textbf{0.7090} & 61.46 & \textbf{69.33} & \textbf{65.76} & \textbf{38.65} \\
Antonyms & 0.7694 & 0.6417 & 0.2730 & 0.4644 & 0.6973 & 61.64 & 66.66 & 64.38 & 38.01 \\
Hypernyms & 0.7759 & 0.6713 & 0.2638 & 0.4554 & 0.6987 & 61.22 & 68.89 & 65.41 & 38.21 \\
Hyponyms & 0.7660 & 0.6324 & 0.2655 & 0.4570 & 0.6972 & 61.38 & 68.28 & 65.15 & 38.30 \\
Member-holonyms & 0.7681 & 0.6321 & 0.2743 & 0.4604 & 0.6952 & \textbf{61.69} & 66.36 & 64.24 & 37.95 \\
Member-meronyms & 0.7701 & 0.6223 & 0.2739 & 0.4611 & 0.6963 & 61.61 & 66.31 & 64.17 & 37.98 \\
Part-holonyms & 0.7852 & 0.6841 & 0.2732 & 0.4650 & 0.7007 & 61.44 & 67.34 & 64.66 & 38.07 \\
Part-meronyms & 0.7786 & 0.6691 & \textbf{0.2761} & 0.4679 & 0.7005 & 61.66 & 67.11 & 64.63 & 38.29 \\
 \hline
\end{tabular}
\vspace{-3mm}
\end{table*}

For the word analogy prediction task we used two benchmarks: Google word analogy dataset~\citep{Mikolov:NIPS:2013},
and SemEval 2012 Task 2 dataset~\citep{SemEavl2012:Task2} (\textbf{SemEval}). 
Google dataset consists of $10,675$ syntactic (\textbf{syn}) analogies, and $8869$ semantic analogies (\textbf{sem}).
The SemEval dataset contains manually ranked word-pairs
for $79$ word-pairs describing various semantic relation types, such as \emph{defective}, and \emph{agent-goal}. 
In total there are $3218$ word-pairs in the SemEval dataset.
Given a proportional analogy  $a:b :: c:d$, we compute the cosine similarity between $\vec{b} - \vec{a} + \vec{c}$ and
$\vec{c}$, where the boldface symbols represent the embeddings of the corresponding words. For the Google dataset, we measure the
accuracy for predicting the fourth word $d$ in each proportional analogy from the entire vocabulary.
We use the binomial exact test with Clopper-Pearson confidence interval to test for the statistical significance of
the reported accuracy values. 
For SemEval we use the official evaluation tool\footnote{\url{https://sites.google.com/site/semeval2012task2/}}
to compute MaxDiff scores.

In Table~\ref{tbl:overall}, we compare the word embeddings learnt by the proposed method for different semantic relation
types in the WordNet.
All word embeddings compared in Table~\ref{tbl:overall} are $300$ dimensional. 
We use the \emph{WordSim-353} (\textbf{WS}) dataset~\citep{Finklestein:02} 
as validation data to find the optimal value of $\lambda$ for each relation type.
Specifically, we minimize \eqref{eq:overall} for different $\lambda$ values,
and use the learnt word representations to measure the cosine similarity for the word-pairs in the \textbf{WS} dataset.
We then select the value of $\lambda$ that gives the highest Spearman correlation
with the human ratings on the \textbf{WS} dataset. This procedure is repeated separately with each semantic relation type $R$.
We found that $\lambda$ values greater than $10000$ to perform consistently well on all relation types. 
The level of performance if we had used only the corpus for learning word representations (without using a semantic lexicon)
 is shown in Table~\ref{tbl:overall} as the \textbf{corpus only} baseline.
This baseline corresponds to setting $\lambda = 0$ in \eqref{eq:overall}.

From Table~\ref{tbl:overall}, we see that by incorporating most of the semantic relations found in the WordNet we can
improve over the corpus only baseline. 
In particular, the improvements reported by synonymy over the \textbf{corpus only} baseline is statistically significant on \textbf{RG},
\textbf{MC}, \textbf{SCWS}, \textbf{MEN}, \textbf{syn}, and \textbf{SemEval}.
Among the individual semantic relations, synonymy consistently performs well on all benchmarks.
Among the other relations, part-holonyms and member-holonyms perform best respectively for predicting semantic similarity
between rare words (\textbf{RW}), and for predicting semantic analogies (\textbf{sem}) in the Google dataset.
Meronyms and holonyms are particularly effective for predicting semantic similarity between rare words.
This result is important because it shows that a semantic lexicon can assist the 
representation learning of rare words, among which the co-occurrences are small even in large corpora~\citep{Luong:CoNLL:2013},
The fact that the proposed method could significantly improve performance on this task 
empirically justifies our proposal for using a semantic lexicon in the word representation learning process.
Table~\ref{tbl:overall} shows that not all relation types are equally useful for learning word representations for a particular task.
For example, hypernyms and hyponyms report lower scores compared to the corpus only baseline on 
predicting semantic similarity for rare (\textbf{RW}) and ambiguous (\textbf{SCWS}) word-pairs.

\begin{table}[t]
\small
\centering
\caption{Comparison against prior work.}
\label{tbl:prior}
\begin{tabular}{|l|c|c|c|c|}\hline 
Method					&	RG		&	MEN	&	sem		&	syn \\ \hline \hline
RCM					&	0.471	&	0.501	&	-		&	29.9 \\
R-NET					&	-		&	-		&	32.64	&	43.46 \\
C-NET					&	-		&	-		&	37.07	&	40.06 \\
RC-NET					&	-		&	-		&	34.36	&	44.42 \\
Retro (CBOW)			&	0.577	&	0.605	&	36.65	&	52.5 \\
Retro (SG)				&	0.745	&	0.657	&	45.29	&	65.65 \\
Retro (corpus only)		&	0.786	&	0.673	&	61.11	&	68.14 \\
Proposed (synonyms)		&	\textbf{0.787}	&	\textbf{0.709}	&	\textbf{61.46}	&	\textbf{69.33} \\ \hline
\end{tabular}
\end{table}

In Table~\ref{tbl:prior}, we compare the proposed method against previously proposed 
word representation learning methods that use a semantic lexicon:
\textbf{RCM} is the relational constrained model proposed by~\citet{Yu:ACL:2014}, 
\textbf{R-NET}, \textbf{C-NET}, and \textbf{RC-NET} are proposed by~\citet{Xu:2014},
and respectively use relational information, categorical information, 
and their union from the WordNet for learning word representations,
and \textbf{Retro} is the retrofitting method proposed by~\citet{faruqui-EtAl:2015:NAACL-HLT}.
Details of those methods are described in Section~\ref{sec:related}.
For \textbf{Retro}, we use the publicly available implementation\footnote{\url{https://github.com/mfaruqui/retrofitting}}
 by the original authors, 
and use pre-trained word representations on the same ukWaC corpus as used
by the proposed method. Specifically, we retrofit word vectors produced by CBOW (\textbf{Retro (CBOW)}), and
skip-gram (\textbf{Retro (SG)}). Moreover, we retrofit the word vectors learnt by the corpus only baseline
(\textbf{Retro (corpus only)}) to compare the proposed \emph{joint} learning approach to the \emph{post-processing} approach
in retrofitting. Unfortunately, for \textbf{RCM}, \textbf{R-NET}, \textbf{C-NET}, and \textbf{RC-NET} 
their implementations, nor trained word vectors were publicly available. Consequently, we report the published results for those methods. 
In cases where the result on a particular benchmark dataset is not reported in the original publication, we have indicated this
by a dash in Table~\ref{tbl:prior}.

Among the different semantic relation types compared in Table~\ref{tbl:overall}, we use the synonym relation which reports
the best performances for the proposed method in the comparison in Table~\ref{tbl:prior}.
All word embeddings compared in Table~\ref{tbl:prior} are $300$ dimensional and use the WordNet as the sentiment lexicon.
From Table~\ref{tbl:prior}, we see that the proposed method reports the best scores on all benchmarks.
Except for the smaller (only 65 word-pairs) \textbf{RG} dataset where the performance of  retrofitting is similar to that of the proposed method,
in all other benchmarks the proposed method statistically significantly outperforms prior work that use a semantic lexicon
for word representation learning.

\begin{figure}[t]
\centering
\includegraphics[width=90mm]{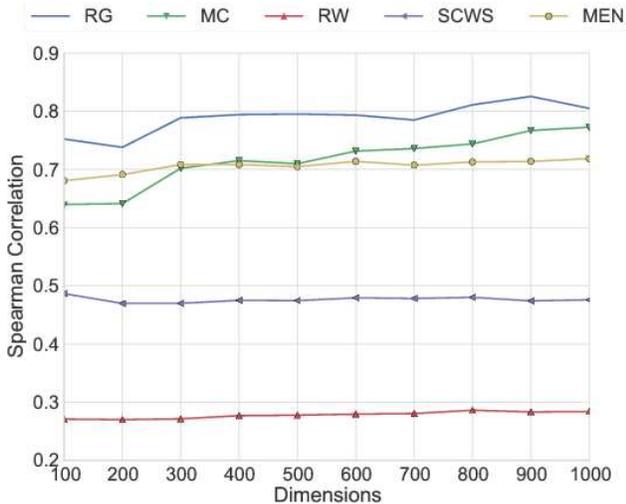}
\caption{The effect of the dimensionality of the word representations learnt by the proposed method using
the synonymy relation, evaluated on semantic similarity prediction task.}
\label{fig:dims}
\end{figure}

We evaluate the effect of the dimensionality $d$ on the word representations learnt by the proposed method.
For the limited availability of space, in Figure~\ref{fig:dims}
we report results when we use the synonymy relation in the proposed method and on the semantic similarity benchmarks. 
Similar trends were observed for the other relation types and benchmarks.
From Figure~\ref{fig:dims} we see that the performance of the proposed method is relatively stable across a wide range of
dimensionalities. In particular, with as less as $100$ dimensions we can obtain a level of performance that outperforms the
corpus only baseline. 
On \textbf{RG}, \textbf{MC}, and \textbf{MEN} datasets we initially see a gradual increase in performance with the dimensionality of
the word representations. However, this improvement saturates after $300$ dimensions, which indicates that it is sufficient to
consider $300$ dimensional word representations in most cases. More importantly, adding new dimensions does not result in
any decrease in performance. 

\begin{figure}[t]
\centering
\includegraphics[width=90mm]{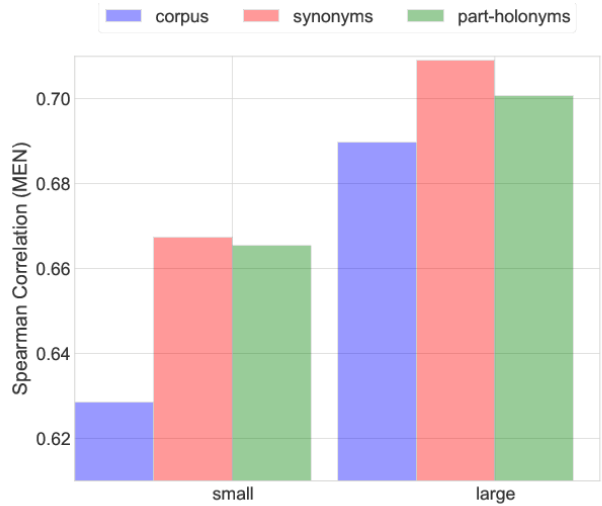}
\caption{The effect of using a semantic lexicon under different corpus sizes. The performance gain is higher when the corpus size is small.}
\label{fig:size}
\end{figure}

To evaluate the effect of the corpus size on the performance of the proposed method, we select a random
subset containing $10\%$ of the sentences in the ukWaC corpus, which we call the \emph{small} corpus, as opposed to
the original \emph{large} corpus. In Figure~\ref{fig:size}, we compare three settings:
\textbf{corpus} (corresponds to the baseline method for learning using only the corpus, without the semantic lexicon),
\textbf{synonyms} (proposed method with synonym relation), and \textbf{part-holonyms} (proposed method with part-holonym
relation). Figure~\ref{fig:size} shows the Spearman correlation coefficient on the \textbf{MEN} dataset for the semantic similarity
prediction task. We see that in both small and large corpora settings we can improve upon the corpus only baseline
by incorporating semantic relations from the WordNet. In particular, the improvement over the corpus only baseline is
more prominent for the smaller corpus than the larger one. Similar trends were observed for the other relation types as well.
This shows that when the size of the corpus is small, word representation learning methods
can indeed benefit from a semantic lexicon.

\section{Conclusion}

We proposed a method for using the information available in a semantic lexicon to improve the word representations
learnt from a corpus. For this purpose, we proposed a global word co-occurrence prediction method using the semantic relations
in the lexicon as a regularizer. Experiments using ukWaC as the corpus and WordNet as the semantic lexicon show that
we can significantly improve word representations learnt using only the corpus by incorporating the information from the semantic lexicon.
Moreover, the proposed method significantly outperforms previously proposed methods for learning word representations
using both a corpus and a semantic lexicon in both a semantic similarity prediction task, and a word analogy detection task.
The effectiveness of the semantic lexicon is prominent when the corpus size is small. Moreover, the performance
of the proposed method is stable over a wide-range of dimensionalities of word representations.
In future, we plan to apply the word representations learnt by the proposed method in downstream NLP applications
to conduct extrinsic evaluations.

\bibliographystyle{named}
\bibliography{ontoreps}

\end{document}